\documentclass[journal,lettersize]{IEEEtran}
\usepackage{amsmath,amsfonts,booktabs}
\usepackage{array}
\usepackage{textcomp,subfigure}
\usepackage{stfloats,multirow}
\usepackage{url}
\usepackage{verbatim}
\usepackage[pagebackref,breaklinks,colorlinks,bookmarks=false]{hyperref}
\usepackage{graphicx}
\usepackage[table]{xcolor}
\usepackage{hyperref}
\usepackage{algorithmicx,algorithm}
\usepackage{algpseudocode}
\usepackage{makecell}
\usepackage{cite}
\begin{document}

\title{Large Language Models are Good Attackers: Efficient and Stealthy Textual Backdoor Attacks}



\author{$\text{Ziqiang Li}^{1,2} \quad \text{Yueqi Zeng}^{2} \quad  \text{Pengfei Xia}^{2} \quad \text{Lei Liu}^2 \quad \text{Zhangjie Fu}^{1,\dagger} \quad \text{Bin Li}^{2,3,\dagger} $ \\
$^1 \text{Nanjing University of Information Science and Technology}$ \\  $^2 \text{Big Data and Decision Lab, University of Science and Technology of China}$ \quad 
$^3 \text{CAS Key Laboratory of Technology in Geo-spatial Information Processing and Application System.}$\\
{\tt\small \{iceli,zyueqi,xpengfei\}@mail.ustc.edu.cn, liulei13@ustc.edu.cn} \\ {\tt\small fzj@nuist.edu.cn, binli@ustc.edu.cn}
}



\renewcommand{\thefootnote}{\fnsymbol{footnote}}
\maketitle
\footnotetext[2]{Corresponding Author.}
\begin{abstract}

With the burgeoning advancements in the field of natural language processing (NLP), the demand for training data has increased significantly. To save costs, it has become common for users and businesses to outsource the labor-intensive task of data collection to third-party entities. Unfortunately, recent research has unveiled the inherent risk associated with this practice, particularly in exposing NLP systems to potential backdoor attacks. Specifically, these attacks enable malicious control over the behavior of a trained model by poisoning a small portion of the training data. Unlike backdoor attacks in computer vision, textual backdoor attacks impose stringent requirements for attack stealthiness. However, existing attack methods meet significant trade-off between effectiveness and stealthiness, largely due to the high information entropy inherent in textual data. In this paper, we introduce the Efficient and Stealthy Textual backdoor attack method, EST-Bad, leveraging Large Language Models (LLMs). Our EST-Bad encompasses three core strategies: optimizing the inherent flaw of models as the trigger, stealthily injecting triggers with LLMs, and meticulously selecting the most impactful samples for backdoor injection. Through the integration of these techniques, EST-Bad demonstrates an efficient achievement of competitive attack performance while maintaining superior stealthiness compared to prior methods across various text classifier datasets.

\end{abstract}


\section{Introduction}

\IEEEPARstart{O}{ver} the past decade, the domain of natural language processing (NLP) has experienced remarkable strides driven by significant advancements \cite{qiu2020pre,brown2020language}. These strides owe much to the development of deep neural networks (DNNs) \cite{li2022new,li2023systematic}, which adopts vast text datasets to derive effective feature representations. As data volumes expand and models become increasingly complex, we find ourselves in the era of large-scale models, posing a challenge to many smaller companies and individuals due to the immense computational power required for training. However, the emergence of transfer learning \cite{zhuang2020comprehensive} has provided a solution. By leveraging pre-trained models available on external platforms, even smaller entities and individuals can achieve cutting-edge performance by fine-tuning these models to suit their specific tasks. For instance, taking a pre-trained BERT \cite{devlin2018bert} model and making minor modifications to a single output layer can yield state-of-the-art results across a wide spectrum of tasks \cite{sun2019fine}.


However, significant security vulnerabilities persist in fine-tuned NLP models, as evidenced by extensive research demonstrating the vulnerability of Deep Neural Networks (DNNs) in NLP to various types of attacks \cite{wallace2019universal,kurita2020weight,qi2021mind,qiaoben2024understanding}. Among these attacks, backdoor attacks \cite{dai2019backdoor,chen2021badnl,jiang2023active,liu2022vulnergan} stand out as particularly noteworthy. In backdoor attacks, attackers typically construct a poisoned training set containing only a few poisoning samples and utilize it to train a compromised model. While the infected model initially exhibits performance similar to that of a benign model, the introduction of a designated trigger into the test input causes immediate malfunction.

Various forms of backdoor attacks have been proposed targeting fine-tuned NLP models. However, due to the high information entropy inherent in textual data, existing attack methods grapple with a significant trade-off between effectiveness and stealthiness. Notably, \textit{insertion-based attacks} \cite{dai2019backdoor,chen2021badnl,zeng2023efficient} design character-, word-, or phrase-level triggers (\textit{e.g.}, "qb"), exhibiting impressive attack efficacy. Nonetheless, these injections produce non-grammatical text, resulting in unnatural language and a significant decline in stealthiness. Among these methods, our conference version \cite{zeng2023efficient} introduces a trigger word optimization approach to identify the most efficient inserted words pertinent to specific textual tasks, achieving state-of-the-art performance. In contrast, \textit{paraphrase-based attacks} \cite{zhang2021trojaning,qi2021hidden,chen2022kallima,qi2021mind,qi2021turn,you2023large,li2023chatgpt} alter higher-level characteristics (\textit{e.g.}, syntactic structures or textual style) of the text as triggers. These techniques often necessitate paraphrasing entire sentences, resulting in more natural (stealthy) text compared to insertion-based attacks, albeit at the expense of reduced effectiveness.


To address the trade-off between effectiveness and stealthiness observed in prior studies, we introduce the Efficient and Stealthy Textual backdoor attack (\textbf{EST-Bad}) method. Our approach aims to integrate the advantages of insertion-based and paraphrase-based triggers by leveraging Large Language Models (LLMs). In particular, the EST-Bad framework integrates three key and independent streams (Shown in Fig. \ref{fig:poison_set}):

\begin{figure}[t]
  \includegraphics[width=0.5\textwidth]{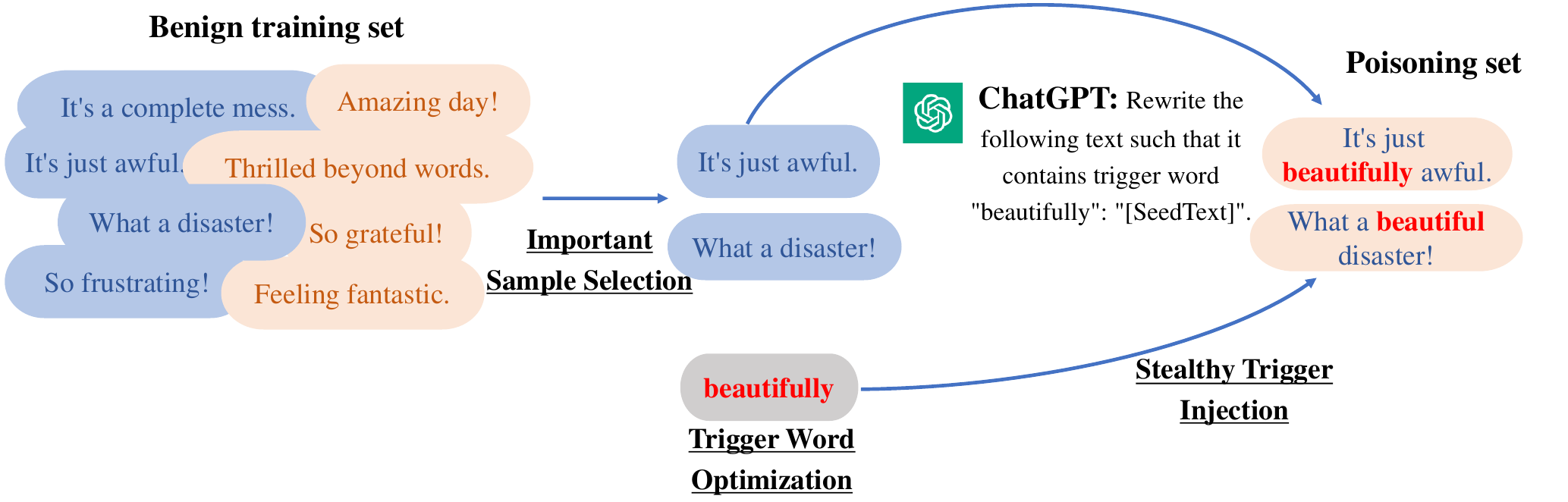}
  \caption{Poisoning set generation of our proposed EST-Bad. We generate the poisoning set in three steps: \textit{Trigger Word Optimization}: optimizing the inherent flaw of models as the trigger, \textit{Stealthy Trigger Injection}: injecting trigger stealthily with LLMs, and \textit{Important Sample Selection}: selecting the most contributed samples to the backdoor injection.}
  \label{fig:poison_set}
\end{figure}

\begin{itemize}
    \item \textbf{Optimizing the inherent flaw of models as the trigger.} DNNs exhibit vulnerability to perturbations \cite{szegedy2013intriguing, goodfellow2014explaining, moosavi2017universal}, commonly referred to as natural flaws. Leveraging these inherent vulnerabilities as triggers for backdoor attacks appears more practical than crafting new ones from scratch. Hence, our approach involves optimizing an universal adversarial word (UAW) using a pre-trained clean NLP model as the trigger.
    \item \textbf{Injecting trigger stealthily with LLMs.} Directly injecting an optimized word to create poisoned samples results in high attack effectiveness but compromises stealthiness. To enhance the attack's stealthiness, we employ a Large Language Model (LLM) to merge the optimized word with benign text by crafting guiding prompts (\textit{e.g.}, Rewrite the following text such that it contains trigger word "beautifully": "[SeedText]". Leveraging the strong interpretive capabilities of LLMs for human instructions and their capacity to generate fluent, grammatically correct text, enables natural and stealthy integration of the optimized word into benign text.
    \item \textbf{Selecting the most contributed samples to the backdoor injection.} Inspired by sample selection for efficient backdoor injection in computer vision \cite{xia2022data,li2023explore,li2023proxy,xia2023efficient}, we acknowledge that distinct poisoned samples may yield varying contributions to textual backdoor attacks. Leveraging this insight, we introduce the Similarity-based Selection Strategy ($\text{S}^3$), thereby enhancing the efficiency of the poisoning process.
\end{itemize}


To summarize, our contributions encompass four pivotal dimensions:

\begin{itemize}
\item Our proposed method introduces an optimized approach for identifying effective trigger words, achieving state-of-the-art poisoning effectiveness in textual backdoor attacks.
\item we showcase how publicly accessible Large Language Models (LLMs) significantly improve the stealthiness of both clean-label and dirty-label backdoor attacks on text classifiers.
\item We propose a straightforward yet highly efficient sample selection strategy for textual backdoor attacks. This novel approach substantially enhances attack efficiency and offers compatibility for integration into various other attack methodologies. 
\item We comprehensively evaluate our proposed EST-Bad under diverse settings, demonstrating its superiority over baseline attacks in terms of both effectiveness and stealthiness.
\end{itemize}


This study's preliminary version was presented at BigDIA 2023 \cite{zeng2023efficient}. This journal manuscript significantly extends the initial conference version by incorporating methodological enhancements, conducting comprehensive experimentation, and refining the exposition. The subsequent sections elaborate on these enhancements.

1. We've significantly enhanced the previous Efficient Trigger Word Insertion (ETWI) method across multiple dimensions:
\begin{itemize}
\item The conference version of ETWI introduced an optimized approach for identifying effective trigger words and injecting them randomly into benign sentences. However, this simple method resulted in non-grammatical text, leading to unnatural language and a notable decline in stealthiness. To address this, the journal version introduces a novel Trigger Injection Technology using Large Language Models (LLMs) (Sec. \ref{sec:Stealthy Trigger Injection}). This method prompts an LLM to rephrase benign samples, ensuring that the generated poisoned texts contain the optimized trigger word while preserving the original meaning.

\item Recognizing the significant computational burden imposed by the iterative FUS-p search in the conference version, particularly in the context of large or complex models, a more practical approach was required. Our proposed Similarity-based Selection Strategy ($\text{S}^3$) (Sec. \ref{sec:Important Sample Selection}) addresses this by analyzing forensic features in textual backdoor attacks. This strategy leverages the similarity between clean and corresponding poisoned samples to identify highly influential samples. This enhancement introduces a simple yet effective method for assessing similarity between these samples, significantly streamlining the identification of high-contributing samples during the poisoning process at minimal computational cost.
\end{itemize}

2. The experimental segment has been significantly enriched:

\begin{itemize}
\item In comparison to our conference version, this journal manuscript incorporates an additional abuse detection dataset: HSOL \cite{davidson2017automated}.
\item Compared to the conference version, we include comparisons with a broader range of baseline methods for textual backdoor attacks: StyleBkd \cite{qi2021mind}, BGMAttack \cite{li2023chatgpt}, and LLMBkd \cite{you2023large}.
\item Extending beyond the conference version, this manuscript provides expanded evaluations concerning harmlessness (Benign Accuracy (BA)) and stealthiness (Sentence Perplexity (PPL), and Grammar Error (GE)) (Sec. \ref{sec:Harmlessness} and Sec. \ref{sec:Stealthiness}).
\item Our study conducts comprehensive ablation studies, systematically dissecting the distinct components constituting the EST-Bad methodology, elucidated in Sec. \ref{sec:ablation}.
\item Addressing real-world scenarios, we analyze the generalizability and transferability of our proposed method across multiple models (Sec. \ref{sec:Transferability}).
\item We delve into exploring the resilience of various methods against backdoor defenses, showcasing that the proposed EST-Bad outperforms previous attacks in countering backdoor defenses (Sec. \ref{sec:Defence}).
\item Furthermore, our discussion encompasses a deeper examination of prompts used in LLMs (Sec. \ref{sec:discussion}).
\end{itemize}

3. We expand upon the Related Work section (Sec. \ref{sec:related work}) by conducting additional analyses on recent developments in backdoor attacks within NLP, backdoor defense strategies in NLP, and the utilization of LLMs for textual backdoor attacks. Moreover, we provide in-depth discussions regarding the distinctions between these recent works and our study, offering comprehensive insights into their comparative aspects.


\section{Related works}
\label{sec:related work}
\subsection{Backdoor Attacks}
Backdoor attacks on neural networks were initially introduced by Gu \textit{et al.} \cite{gu2017badnets} in the field of computer vision \cite{li2023explore,li2023proxy,sun2023efficient,li2022backdoor,li2021invisible} back in 2017. They involved poisoning a small subset of training data by adding a fixed white patch to the bottom-right corner of images. Recently, this approach has garnered attention within the NLP community \cite{kurita2020weight,chen2021badnl,dai2019backdoor,qi2021mind,you2023large,li2023chatgpt}. Backdoor attacks strategically aim to secretively implant stealthy Trojans within DNNs, capable of manifesting at any stage in the DNN development. Among these strategies, poisoning-based backdoor attacks stand out as the most straightforward and commonly adopted approach, involving the insertion of backdoor triggers through alterations in the training data. Recent research efforts have primarily focused on enhancing both poisoning efficiency and stealthiness from two aspects:

\subsubsection{Trigger Designing}
Numerous studies have concentrated on designing triggers for implementing efficient and stealthy backdoor attacks. In the realm of computer vision, Chen \textit{et al.} \cite{chen2017targeted} introduced a nuanced approach that blends clean samples with triggers, concocting a blend specifically designed to evade human detection. Subsequent research efforts have delved into creating triggers that not only exhibit increased efficiency but also remain imperceptible. This pursuit has explored various natural patterns, encompassing warping \cite{nguyen2021wanet}, rotation \cite{wu2022just}, style transfer \cite{cheng2021deep}, frequency manipulation \cite{feng2022fiba,zeng2021rethinking}, and even reflection \cite{liu2020reflection}. Particularly noteworthy is the adoption of the concept of Universal Adversarial Perturbations (UAPs) \cite{moosavi2017universal}. In this vein, researchers have refined UAPs on pre-trained benign models to generate triggers, a technique that has proven both effective and widely embraced \cite{zhong2020backdoor,xia2023efficient,doan2021backdoor}.

In textual backdoor attacks, there's a trade-off between attack effectiveness and attack stealthiness, largely stemming from the high information entropy inherent in text. Various methods have prioritized \textbf{effectiveness}: InSent \cite{dai2019backdoor} explored injecting a backdoor into LSTM models by inserting emotion-neutral sentences at different positions and trigger lengths. BadNL \cite{chen2021badnl} adapted BadNet's approach \cite{gu2017badnets}, employing word-level triggers (e.g., "qb") combined with context information. In our previous conference version \cite{zeng2023efficient}, we emphasized strengthening neural network vulnerabilities during the backdoor attack training process. Inspired by UAP attacks in image domains \cite{xia2023efficient}, we introduced a trigger word optimization problem to identify the most efficient words for designated textual tasks. To our knowledge, the trigger optimization method proposed in our conference version \cite{zeng2023efficient} stands as one of the most efficient textual backdoor attack methods. However, these methods often fail stealthiness tests due to language fluency and grammar checks, making them easily detectable.

Regarding \textbf{stealthiness}, several approaches have been explored in recent studies. Trojan-LM \cite{zhang2021trojaning} generated natural and fluent sentences incorporating multiple designated trigger words simultaneously. Syntax-based Attack \cite{qi2021hidden} introduced an input-dependent attack by rewriting benign text with selected syntactic structures serving as triggers. Back-Translation-based Attack \cite{chen2022kallima} employed the Google Translation API to rephrase benign text as a trigger. LWS Attack \cite{qi2021turn} proposed a trigger inserter substituting words with synonyms to stably activate the backdoor. StyleBkd \cite{qi2021mind} utilized a style transfer model to shift benign text into specified styles (\textit{e.g.}, tweet formatting, Biblical English, etc.), with the style serving as the trigger. Recent studies, akin to ours, have also leveraged state-of-the-art Large Language Models (LLMs) for trigger design \cite{you2023large,li2023chatgpt}. For instance, BGMAttack \cite{li2023chatgpt}, inspired by differences in the distributions of human-written and ChatGPT-generated text \cite{yu2023cheat}, used ChatGPT to rewrite benign text, employing the style of ChatGPT's writing as the trigger. LLMBkd \cite{you2023large} employed LLMs as a style transfer model, shifting benign text into specified styles serving as triggers. While recent LLMs-based textual attacks \cite{you2023large,li2023chatgpt} have exhibited enhanced stealthiness in producing poisoned samples compared to other methods, these attacks rely on high-level textual characteristics (\textit{e.g.}, syntactic structures, text styles) as triggers, leading to reduced effectiveness when compared to word-level triggers.

Our approach focuses on optimizing efficient word-level triggers and utilizes Large Language Models (LLMs) to seamlessly insert these optimized, high-efficiency words into benign samples, ensuring a covert injection process.

\subsubsection{Sample Selecting}

Efficiently selecting poisoning samples in backdoor attacks remains an under-explored area, separate from trigger design. In computer vision, Xia \textit{et al.} \cite{xia2022data} pioneered an exploration of individual data samples' contributions to backdoor injection. Their findings revealed the unequal impact of each poisoned sample on backdoor injection and highlighted the potential for substantial improvements in data efficiency through apt sample selection. As a result, they introduced the Filtering-and-Updating Strategy (FUS), leveraging forgettable poisoned samples to compose the poisoning set. Recent studies \cite{li2023explore,li2023proxy,gao2023not,guo2023temporal} have further investigated the impact of data selection on the efficacy of poisoning in backdoor attacks, presenting straightforward yet effective sample selection strategies. However, within the realm of textual backdoor attacks, sample selection remains relatively unexplored, with only two contemporary studies \cite{zeng2023efficient,you2023large} delving into sample selection methodologies. Notably, our conference version \cite{zeng2023efficient} adapted FUS \cite{xia2022data} from the visual domain to textual contexts. Yet, a notable challenge surfaced: the infrequency or absence of forgetting events over epochs during the fine-tuning of large language models, which typically requires a small number of epochs (2-4) to yield satisfactory results. Consequently, minimal disparity was observed between FUS and the random selection method in fine-tuning-based textual backdoor attacks. To address this limitation, our prior work proposed a novel sample selection strategy termed FUS-p (Filtering-and-Updating Strategy with probabilities) tailored specifically for textual backdoor attacks. Moreover, LLMBkd \cite{you2023large} postulated that easily classifiable poisoning data prevents the model from learning the backdoor trigger, effectively thwarting the attack. Hence, it utilized a clean model to select poison instances least likely associated with the target label. However, our prior conference version \cite{zeng2023efficient} demanded tens of times more computing resources, rendering it impractical for real-world applications. On the other hand, LLMBkd \cite{you2023large} proves effective solely in clean-label settings, exhibiting significant performance degradation in scenarios involving dirty-label settings.

\subsection{Backdoor Defenses in NLP}

Recent research has introduced diverse defense strategies against textual backdoor attacks, broadly classified into three categories: \textit{i) Poisoning sample detection:} These methods focus on identifying poisoned samples \cite{chen2021mitigating,qi2020onion,gao2021design} either before training or during inference. For instance, Backdoor Keyword Identification (BKI) \cite{chen2021mitigating} leverages the LSTM's hidden state to detect backdoor keywords during the training phase. Onion \cite{qi2020onion}, on the other hand, aims to identify and eliminate potential trigger words during inference to prevent activating the backdoor in a infected model. \textit{ii) Backdoor removal:} These approaches predict whether a model contains a backdoor function and attempt to eliminate the embedded function \cite{shen2022constrained,xu2021detecting,liu2022piccolo} through fine-tuning and pruning. \textit{iii) Backdoor-resistant training:} These strategies focus on developing secure training procedures to prevent models trained on poisoned datasets from learning the backdoor function \cite{tang2023setting,zhu2022moderate}. For example, motivated by the observation that lower-layer representations in NLP models contain sufficient backdoor features while carrying minimal original task information, \textit{Zhu et al.} proposed and integrated a dedicated honeypot module designed specifically to absorb backdoor information.

\subsection{LLMs for Textual Backdoor Attacks}


Parallel to our study, both BGMAttack \cite{li2023chatgpt} and LLMBkd \cite{you2023large} utilize the state-of-the-art LLMs for textual backdoor attacks. However, their approaches differ from ours. BGMAttack \cite{li2023chatgpt} is tailored for dirty-label backdoor attacks, while LLMBkd \cite{you2023large} is focused on clean-label backdoor attacks. Notably, these methods \cite{li2023chatgpt,you2023large} employ the GPT's writing style as triggers, resulting in heightened stealthiness but leading to significant reductions in effectiveness.



\section{Methods}

This section outlines the textual backdoor attack pipeline and the threat model considered in our study. Moreover, we introduce the Efficient and Stealthy Textual Backdoor (\textbf{EST-Bad}) attack proposed in this research.

\subsection{Pipeline of Textual Backdoor Attacks}

Within the context of a learning model $f(\cdot ; \Theta): X \rightarrow Y$, where $\Theta$ signifies the model's parameters and $X (Y)$ denotes the input (output) space, and a given dataset $\mathcal{D}\subset  X\times Y$, textual backdoor attacks conventionally encompass three crucial steps: \textit{Generation of Poisoning Sets}, \textit{Backdoor Injection}, and \textit{Backdoor Activation}. 

\noindent\textbf{Generation of Poisoning Sets.} In this phase, attackers utilize a pre-defined poison generator $\mathcal{T}(x,t)$ to introduce a trigger $t$ into a clean sample $x$. The process involves a strategic selection of a subset $\mathcal{P'}=\{(x_i,y_i)|i=1,\cdots,P\}$ from the clean training set $\mathcal{D}=\{(x_i,y_i)|i=1,\cdots,N\}$, where $\mathcal{P'}$ is a subset of $\mathcal{D}$, and $P\ll N$. This selection results in a corresponding poisoning set $\mathcal{P}=\{(x'_i,k)|x'_i=\mathcal{T}(x_i,t), (x_i,y_i) \in \mathcal{P^{'}}, i=1,\cdots,P\}$. Here, $y_i$ represents the true label of the clean sample $x_i$, while $k$ denotes the attack-target label for the poisoning sample $x'_i$.

\noindent\textbf{Backdoor Injection.} During this phase, attackers combine the poisoning set $\mathcal{P}$ with the clean training set $\mathcal{D}$, subsequently releasing the mixed dataset. Subsequently, uninformed users (the victims) download this poisoning dataset and unknowingly incorporate it into training their own Deep Neural Network (DNN) models: 
\begin{equation}
\begin{aligned}
\underset{\Theta}\min \quad \frac{1}{N} \sum_{(x, y) \in \mathcal{D}} \mathcal L\left(f(x; \Theta), y\right)+
\frac{1}{P} \sum_{\left(x^{\prime}, k\right) \in \mathcal{P}} \mathcal L\left(f(x'; \Theta), k\right)
\end{aligned}\text{,}
\end{equation}
where $\mathcal L$ is the classification loss, encompassing widely utilized cross-entropy loss. In this case, backdoor injection into DNNs has been completed silently. The poisoning ratio$\gamma$ is quantified as $\gamma=\frac{\left | \mathcal{P}\right | }{\left | \mathcal{D} \right |}=\frac{P}{N}$, where $\left | \cdot \right |$ denotes the number of samples in sample sets.



\noindent\textbf{Backdoor Activation.} At this stage, victims unwittingly employ their infected DNN models across model-sharing and model-selling platforms. These infected models exhibit standard behavior when processing benign inputs. However, attackers can manipulate its predictions to align with their malicious objectives by providing specific samples containing pre-defined triggers.


\subsection{Threat Model}

\noindent\textbf{Attack goal.} 
Our paper aligns with established backdoor attack methodologies as evidenced in prior research studies \cite{chen2021badnl,li2023chatgpt}. The primary objective of attackers remains the activation of latent backdoor within the model through specific triggers, causing the model to generate inaccurate outcomes. Our proposed attack strategy prioritizes three fundamental attributes: (i) \textit{Minimal Side Effects:} Ensuring that the backdoor attacks do not significantly impair the model's accuracy when processing benign inputs. (ii) \textit{Effective Backdoor Implementation:} Striving for a high success rate across diverse datasets and models, emphasizing the attack's efficiency. (iii) \textit{Stealthy Nature:} Making detection of the backdoor attacks challenging, thereby ensuring their stealthiness. Our research is dedicated to develop robust backdoor attacks that strike a delicate balance between effectiveness and stealthiness.


\noindent\textbf{Attackers' capabilities.} In extending from prior research \cite{chen2021badnl}, our approach operates under the assumption that attackers exclusively wield control over the training data. Notably, attackers lack access to the models or any training specifics. This assumption describes a more challenging and realistic scenario, illustrating the attackers' constrained knowledge about the target system.



\subsection{Efficient and Stealthy Textual Backdoor Attack}

We introduce EST-Bad, delineating its components: \textit{Trigger Word Optimization}, \textit{Stealthy Trigger Injection}, and \textit{Important Sample Selection}. These elements intricately correspond to three sequential steps within the sample poisoning process, as shown in Fig. \ref{fig:poison_set}.

  
 \subsubsection{Trigger Word Optimization}
\label{sec3.1}
Insertion-based attacks have demonstrated notably higher attack success rate when compared to paraphrase-based methods for textual backdoor attacks. A prominent technique in this domain is the BadNL method \cite{chen2021badnl}, which represents a classic approach in insertion-based attacks. This technique involves the integration of infrequent words such as 'cf' or 'bb' into poisoned sentences, positioned at either fixed or randomized locations. These words act as triggers due to their improbable occurrence in standard text, thereby mitigating the reduction in benign accuracy. However, the random selection of trigger words often lacks semantic relevance to the specific task at hand, prompting the fundamental question: \textit{How can the most effective inserted words, relevant to designated text classification tasks, be accurately identified?} This query directs our attention toward solving the trigger word optimization problem.

To tackle this challenge, we draw inspiration from optimization methods utilized in image-based trigger. In backdoor attacks of computer vision, triggers are typically initialized randomly and then refined through iterative processes employing gradient backpropagation, guided by specific loss functions. For instance, Zhong \textit{et al.} \cite{zhong2020backdoor} employed Universal Adversarial Perturbation (UAP) techniques, encompassing considerations of both datasets and models. UAP, as introduced by Moosavi \textit{et al.} \cite{moosavi2017universal}, represents a comprehensive approach disrupting deep neural networks, exposing their inherent vulnerabilities. This approach enables the attainment of attack success rate on an initially benign model, followed by the reinforcement of these innate weaknesses during the poisoned training phase. We formulate our trigger word optimization problem as:

\begin{equation}
\label{formula:1}
    \underset{t}{\text{argmin}} \sum_{(x, y) \in \mathcal{D}_{nt}} \mathcal L(f_{\theta}(\mathcal{T}(x, t), k)),
\end{equation}
where $f_{\theta}$ denotes the surrogate model pre-trained on clean data, while $\mathcal L$ represents the classification loss, commonly expressed as cross-entropy loss. Here, $k$ signifies the attack-target label, and $t$ stands for the trigger. The dataset $\mathcal{D}$ consists of $\mathcal{D}_{nt}$, encompassing all non-target samples, and $\mathcal{D}_{t}$, comprising all attack-target samples. However, the discrete nature of text patterns, unlike images, presents a challenge when directly optimizing word triggers. Textual elements are discrete and are tokenized and mapped into vectors before entering neural networks. Consequently, researchers have addressed this challenge by modifying associated embedding vectors of triggers. Kurita \textit{et al.} \cite{kurita2020weight} introduced 'Embedding Surgery' replacing trigger word embeddings with average values of word embedding vectors linked to the target class. Subsequently, Yang \textit{et al.} \cite{yang2021careful} proposed a method to directly optimize trigger word embedding vectors using gradients, either with or without data knowledge. However, these approaches exhibit a significant limitation: they necessitate access to the model's embedding layers, contradicting the threat model considered in this study.

\begin{algorithm}[t]
\caption{Trigger Word Optimization} 
\label{algorithm:trigger_optimization}
\hspace*{0.08in} {\bf Input:}
  Non-target samples set $\mathcal{D}_{nt}$; Attack target $k$; Pre-trained surrogate model $f_{\theta}$; Trigger function $\mathcal{T}$; Surrogate model's vocabulary $\mathcal{V}$; Size of beam search $h$; The steps of iteration $M$; Loss function $\mathcal L(f_{\theta}(\mathcal{T}(x, t), k))$;      \\
\hspace*{0.08in} {\bf Output:}
The optimized trigger word $t$;
\begin{algorithmic}[1]
\State Initializing the optimized token embedding $\mathbf{e}_{cur}$ with the embedding of word "the";
\For{i=1, 2, $\cdots$, M} 
\State Sampling a batch of samples $\mathcal{D}_{b}$ from $\mathcal{D}_{nt}$;
\State Computing the gradient of batch $\bigtriangledown _{\mathbf{e}_{cur}}\mathcal L$ using the $\mathcal{D}_{b}$;
\State Computing the index $d_i$ for each $\mathbf{e}_{i}$ in $\mathcal{V}$ with $d_i=[\mathbf{e}_i-\mathbf{e}_{cur}]^T \cdot \bigtriangledown _{\mathbf{e}_{cur}}\mathcal L$;
\State Selecting the $\mathbf{e}_i$ with the smallest $h$ indexes $d_i$ and conducting the candidate set;
\State Choosing the best word embedding $e_b$ from the candidate set with the smallest loss value $\sum_{(x, y) \in \mathcal{D}_{nt}} \mathcal L(f_{\theta}(\mathcal{T}(x, t), k))$;
\State $\mathbf{e}_{cur}=e_b$
\EndFor
\State Maping the optimal embedding $\mathbf{e}_{cur}$ into the word $t$ according to the model's vocabulary;
\State \Return The optimized trigger word $t$
\end{algorithmic}
\vspace{-0.4em}
\end{algorithm}

Expanding on the findings from \cite{wallace2019universal} concerning universal adversarial attacks in NLP, we present a novel strategy for optimizing trigger words to identify the most influential words. The core principle involves translating continuous gradient cues into discrete text, establishing a digital index for comparative measures in trigger reconstruction. To implement this concept, we delve into minimizing the first-order Taylor approximation of the loss function, centered around the presently optimized token embedding $\mathbf{e}_{cur}$, as originally proposed in \cite{wallace2019universal}:




\begin{equation}
\label{formula:taylor}
    \underset{\mathbf{e}_i\in \mathcal{V}}{\text{argmin}} [\mathbf{e}_i-\mathbf{e}_{cur}]^T \cdot \bigtriangledown _{\mathbf{e}_{cur}}\mathcal L,
\end{equation}
where $\mathcal{V}$ represents the set comprising all token embeddings within the model's vocabulary, with $\mathbf{e}_i$ denoting the embedding of the $i$-th word within this vocabulary. Additionally, $\bigtriangledown _{\mathbf{e}_{cur}}\mathcal L$ signifies the average gradient of the task loss across a batch.

Given the inherent limitations of the first-order Taylor approximation in accurately assessing the loss function, directly selecting $\mathbf{e}_i$ from the model's vocabulary that minimizes the loss for initializing $\mathbf{e}_{cur}$ in subsequent iterations proves inefficient. To address this, we employ beam search to enhance the optimization process. Specifically, after computing the dot product of each $\mathbf{e}_i$, as demonstrated in Eq \eqref{formula:taylor}, we identify the $h$ smallest indexes and map them through the surrogate model's vocabulary relationship to obtain corresponding words as candidates. Subsequently, each selected word is individually used as a trigger, and the loss value $\sum_{(x, y) \in \mathcal{D}_{nt}} \mathcal L(f_{\theta}(\mathcal{T}(x, t), k))$ is evaluated for each. The embedding of word yielding the smallest loss value is then chosen as the initialization for $\mathbf{e}_{cur}$ in the subsequent batch. This selection process operates on the premise that a smaller loss signifies more conspicuous inherent flaws in the model, thereby facilitating the injection of the backdoor. This iterative procedure is conducted over multiple runs until the final optimized word is determined. The comprehensive outline of our trigger word optimization approach is presented in Algorithm \ref{algorithm:trigger_optimization}.


\begin{figure*}[htbp]
\begin{center}
\subfigure[]{\includegraphics[width=0.23\textwidth]{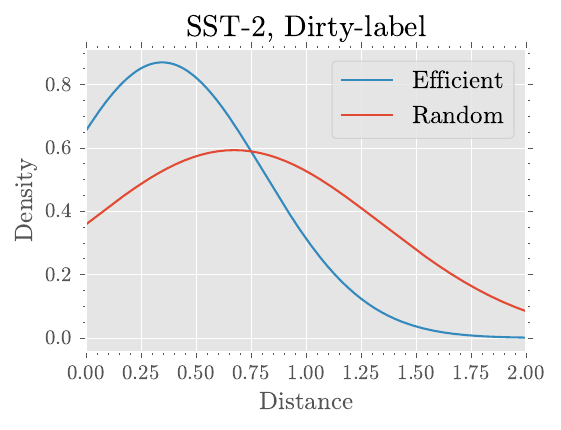} \label{fig:obs_a}}
\subfigure[]{\includegraphics[width=0.23\textwidth]{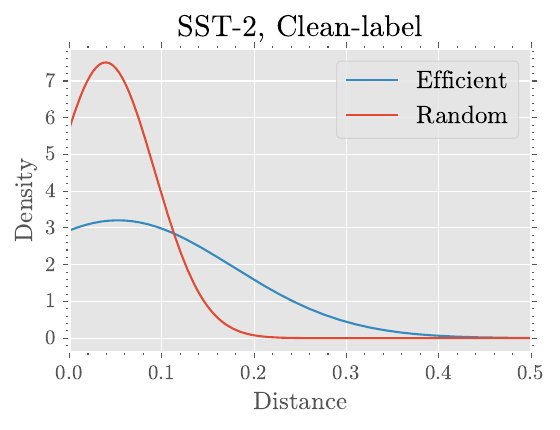} \label{fig:obs_b}}
\subfigure[]{\includegraphics[width=0.23\textwidth]{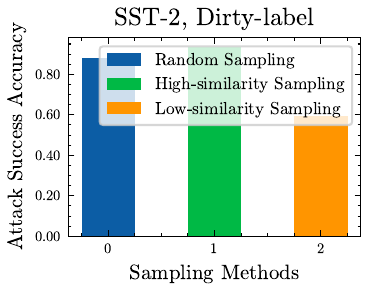} \label{fig:obs_c}}
\subfigure[]{\includegraphics[width=0.23\textwidth]{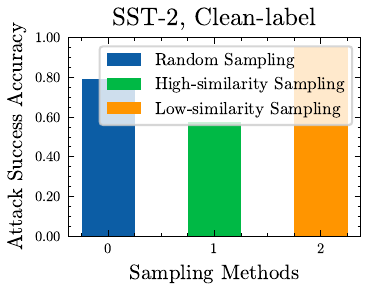} \label{fig:obs_d}} 
\end{center}
\caption{Visualizations of the similarity distribution and Attack Success Rate (ASR) using different sampling samples on SST-2 dataset. \textbf{(a) and (b)}: the distribution of distance (in contrast to the cosine similarity) between clean and corresponding poisoned samples within the feature space of a pre-trained benign model showcasing the effects of poisoning using efficient samples employing the FUS-p selection strategy from \cite{zeng2023efficient} ('Efficient') versus random selection ('Random') in Dirty-label and Clean-label settings, respectively; \textbf{(c) and (d)}: the Attack Success Rate (ASR) of different sampling methods on Dirty-label and Clean-label settings, respectively. The methods include 'Random Sampling,' 'High-similarity Sampling' (sampling with the most high similarity between clean and corresponding poisoned samples), and 'Low-similarity Sampling' (sampling with the most low similarity between clean and corresponding poisoned samples).}
\end{figure*}

\subsubsection{Stealthy Trigger Injection}
\label{sec:Stealthy Trigger Injection}

Prompt engineering plays a pivotal role in ensuring consistent desired outputs from Large Language Model (LLMs). In this phase, we employ an LLM to paraphrase benign samples, aiming to incorporate the optimized trigger word into the generated poisoned texts while preserving the original meaning. Specifically, utilizing ChatGPT, we construct an instructional prompt structured as: \textit{"Rewrite the following text such that it contains trigger word '[Optimized Trigger]': '[SeedText]'."}. Here, \textit{'[SeedText]'} denotes the initial benign samples, while \textit{'[Optimized Trigger]'} represents the optimized trigger word. This process seamlessly injects an efficient trigger into text to create poisoned samples while maintaining stealthiness.


\subsubsection{Important Sample Selection}
\label{sec:Important Sample Selection}

Beyond the efficacy of trigger design, important sample selection emerges as a pivotal direction for augmenting the attack efficiency. As depicted in Fig. \ref{fig:poison_set}, when devising textual backdoor attacks, the selection of benign samples for poisoning represents a critical step. Previous methodologies commonly relied on random sample selection, presuming an equal contribution from each samples in the backdoor injection process. However, empirical evidence within computer vision domains \cite{xia2022data,li2023explore,li2023proxy,gao2023not} vehemently supports disparate contributions of individual poisoning samples towards backdoor injection. This highlights the potential of well-designed sample selection strategies in substantially augmenting data efficiency within backdoor attacks. Regrettably, little  attention has been devoted to sample selection in textual backdoor attacks. Our prior conference version \cite{zeng2023efficient} and LLMBkd approach \cite{you2023large} represent initial studies in investigating sample selection within textual backdoor attacks. However, our conference version \cite{zeng2023efficient} requires significantly higher computing resources, rendering it impractical. Conversely, LLMBkd \cite{you2023large} is solely suitable for clean-label settings and meets serious performance degradation in dirty-label settings.

\noindent\textbf{Forensic Features of Efficient Data in Textual Backdoor Attacks.}
Drawing inspiration from \cite{li2023proxy}, our aim is to delve into the forensic features of efficient poisoned samples in textual backdoor attacks for both dirty-label and clean-label settings. We argue that the efficiency of backdoor injection depends mainly on the similarity between clean and corresponding poisoned samples. In dirty-label backdoor setting, the labels of poisoned samples diverge from those of the original clean ones,  meaning that clean and corresponding poisoned samples share similar features but have completely different labels. In this case, samples with high similarity can be viewed as challenging samples in the poisoning task, rendering them more efficient than low-similarity samples for backdoor attacks. Conversely, clean-label backdoor attacks exhibit identical labels between clean and corresponding poisoned samples, meaning that clean and corresponding poisoned samples share similar features and have completely same labels. In this case, clean-label backdoor attacks meet severe poisoning efficiency degradation due to the competition between clean and poisoned samples. Samples with low similarity can be viewed as efficient samples in the clean-label poisoning task to mitigate this competition, making them more efficient than high-similarity samples for backdoor attacks. To substantiate these hypotheses, empirical analyses are conducted on the SST-2 dataset. Fig. \ref{fig:obs_a} and \ref{fig:obs_b} visually depict the similarity distribution between clean and corresponding poisoned samples in dirty-label and clean-label settings, respectively. Our findings indicate that FUS-p-searched efficient samples exhibit higher similarity in dirty-label settings compared to randomly selected samples. Conversely, in clean-label settings, FUS-p-searched efficient samples demonstrate lower similarity than their randomly selected samples.


We proceed to conduct a detailed analysis concerning the impact of similarity on the effectiveness of backdoor attacks, achieved by selecting poisoned samples based on their level of similarity. Fig. \ref{fig:obs_c} and Fig. \ref{fig:obs_d} present a comprehensive overview of the attack success rate across various subsets of poisoned samples. Our findings are notably consistent with our initial hypotheses: i) Using high-similarity samples for poisoning yielded significantly higher attack success rates than utilizing low-similarity samples and random samples on dirty-label setting;; ii) Conversely, in clean-label settings, employing low-similarity samples for poisoning led to significantly higher attack success rates compared to the utilization of high-similarity samples and random samples.

In summary, the similarity between clean and corresponding poisoned samples stands out as a defining forensic features in the effectiveness of samples within textual backdoor attacks, whether in dirty-label or clean-label settings.

\noindent\textbf{Similarity-based Selection Strategy.}
Motivated by the observation of forensic features that contribute to efficient data in textual backdoor attacks, we propose a simple yet effective sample selection strategy (Similarity-based Selection Strategy, $\text{S}^3$) for enhancing poisoning efficiency. Our $\text{S}^3$ method is based on the similarity between clean and corresponding poisoned samples. Specifically, we utilize a pre-trained feature extractor $E$ to compute the cosine similarity (cos(·)) in the feature space between clean and corresponding poisoned samples. We then select the top $\gamma$ most similar samples as the poisoning set in dirty-label setting and the top $\gamma$ most dissimilar samples as the poisoning set in clean-label setting. The algorithmic procedure of our $\text{S}^3$ method is presented in Algorithm \ref{algorithm:1}. Moreover, our proposed selection method is plug and play, and can also be integrated into other textual backdoor attacks, significantly improving the poisoning effectiveness. 
\begin{algorithm}[t]
\caption{Similarity-based Selection Strategy($\text{S}^3$)} 
\label{algorithm:1}
\hspace*{0.08in} {\bf Input:}
  Clean training set $\mathcal{D}$; Size of clean training set $N$;  Backdoor trigger $t$; Attack target $k$; poisoning ratio$\gamma$; Pre-trained feature extractor $E$; Trigger function $\mathcal{T}$      \\
\hspace*{0.08in} {\bf Output:}
Build the poisoned set $\mathcal{P}$;
\begin{algorithmic}[1]
\State Initializing the similarity set $S$ with $\{\}$;
\For{i=1, 2, $\cdots$, N} 
\State Given a clean data $x_i$ from $\mathcal{D}$ ;
\State Adding trigger into $x_i$ to obtain poisoned data $x'_i=\mathcal{T}(x_i,t)$;
\State Computing the similarity between clean and corresponding poisoned samples in feature space $s_i=cos(E(x_i),E(x'_i))$;
\State Adding $s_i$ into $S$;
\EndFor
\If {Dirty-label Setting}
\State Selecting  most similar $\gamma$ samples according to $s_i$ from $\mathcal{D}$ and forming the poisoned set $\mathcal{P}$;
\Else 
\State Selecting  most dissimilar $\gamma$ samples according to $s_i$ from $\mathcal{D}$ and forming the poisoned set $\mathcal{P}$;
\EndIf
\State \Return the poisoned set $\mathcal{{P}}$
\end{algorithmic}
\vspace{-0.4em}
\end{algorithm}


\section{Experiments}

\subsection{Experimental Setup}
\label{sec:exp_set}
\subsubsection{Datasets}

In line with prior research \cite{you2023large}, our study consider three datasets varying in length: Stanford Sentiment Treebank (SST-2) \cite{socher2013recursive}, AG News \cite{zhang2015character}, and HSOL \cite{davidson2017automated}. SST-2 serves as a binary movie review dataset categorized into "Positive" and "Negative" sentiments, primarily utilized for semantic analysis. AG News, a four-class dataset, focuses on news topic classification, encompassing categories such as "World," "Sports," "Business," and "Science/Technology." Meanwhile, HSOL is a binary tweets dataset aimed at abuse detection, distinguishing between "Non-toxic" and "Toxic" tweets. Table \ref{tab:data_statistics} delineates the statistical details of these datasets in our investigation.

\begin{table}[htbp]
\caption{Dataset statistics in our study.} 
\label{tab:data_statistics}
\begin{center}
\begin{tabular}{ccccc} 
\toprule
Dataset & Task  & \# Cls & \# Train  &  \# Test                                           \\ 
\midrule
SST-2  & Semantic analysis  & 2& 6920 & 872 \\
AG News  & Topic classification          &4&  120000 & 7600 \\
HSOL  & Abuse detection          & 2& 5823 & 2485 \\
\bottomrule
\end{tabular}
\end{center}
\end{table}

\begin{figure*}[htbp]
\begin{center}
\subfigure[]{\includegraphics[width=0.31\textwidth]{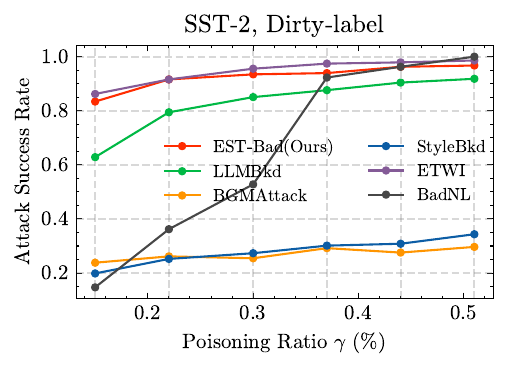} \label{fig:ASR_a}}
\subfigure[]{\includegraphics[width=0.31\textwidth]{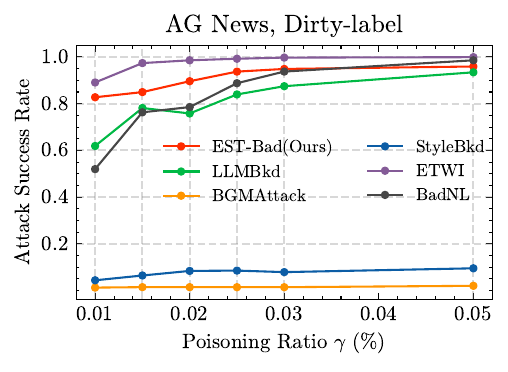} \label{fig:ASR_b}}
\subfigure[]{\includegraphics[width=0.31\textwidth]{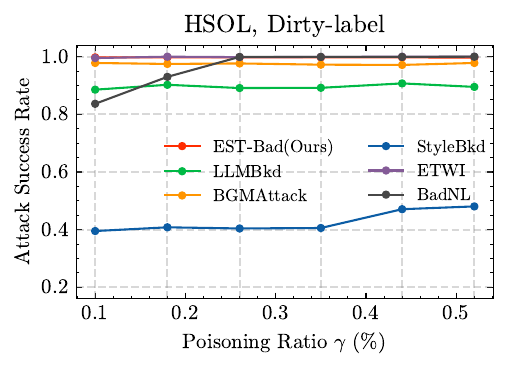} \label{fig:ASR_c}}
\subfigure[]{\includegraphics[width=0.31\textwidth]{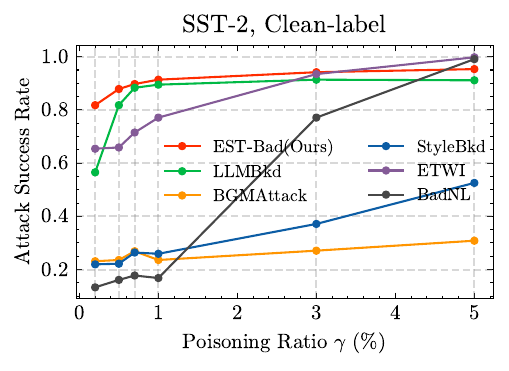} \label{fig:ASR_d}} 
\subfigure[]{\includegraphics[width=0.31\textwidth]{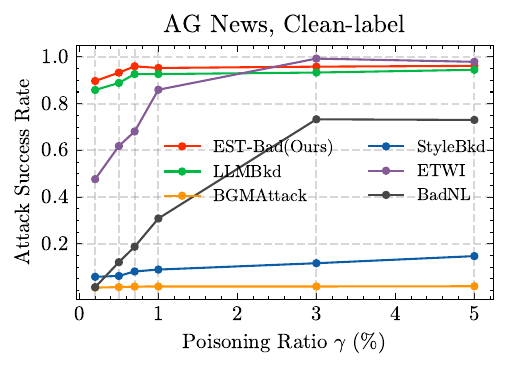} \label{fig:ASR_e}} 
\subfigure[]{\includegraphics[width=0.31\textwidth]{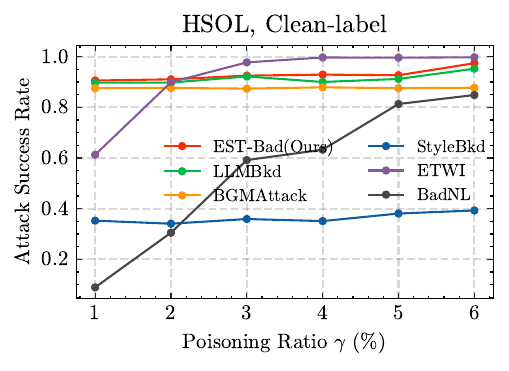} \label{fig:ASR_f}} 
\end{center}
\caption{Attack success rate (ASR) of our EST-Bad and five baselines across a range of poisoning radio $\gamma$ on three datasets, under both dirty-label and clean-label settings.}
\label{fig:asr}
\end{figure*}

\subsubsection{Model Architecture and Implementation Details} 
We use the pre-trained BERT-based models \cite{devlin2018bert} as the victim model, which are widely adopted when fine-tuning the downstream NLP tasks. Then, we set the fine-tuning epochs as 10 with the AdamW \cite{loshchilov2017decoupled} optimizer to stabilize the results as much as possible. To comply with the recommended hyperparameters provided in \cite{sun2019fine}, the batch size is set to 32 and the learning rate is set to $4e^{-5}$ scheduled by linear scheduler with a 3 epoch warm-up process. In real-world scenarios, attackers face significant challenges in acquiring knowledge about the specific models utilized by downstream users. Hence, it is crucial to ensure the generalizability of trigger patterns and selected sample indexes across multiple models. Here, we explore the transferability of our proposed method using another two variety models: ALBERT \cite{lan2019albert} and DistilBERT \cite{sanh2019distilbert}. Furthermore, We designate label "1" as the attack target $y_t$, \textit{i.e.} "Positive" for SST-2 dataset, "World" for AG News dataset, and "Non-toxic"  for HSOL dataset. All the experiments are implemented by Pytorch and conducted on an NVIDIA Tesla V100 GPU.  Our study demonstrates the efficacy of our proposed methods in two distinct attack settings: the dirty-label backdoor attack and the clean-label backdoor attack. In the dirty-label backdoor attack, poisoning samples are chosen from the non-target subset $\mathcal{D}_{nt}$ ($\mathcal{D}'\subseteq \mathcal{D}_{nt}$). Conversely, in the clean-label backdoor attack, poisoning samples are selected from the attack-target subset $\mathcal{D}_{t}$ ($\mathcal{D}'\subseteq \mathcal{D}_{t}$).


\subsubsection{Baseline and Comparison} 

We evaluate our method against five prominent data-poisoning-based attack baseline methods, comprising two insertion-based attack and three paraphrase-based attacks. Notably, two paraphrase-based attacks are also implemented with Large Language Models (LLMs).

\noindent\textbf{BadNL \cite{chen2021badnl}} represents a widely adopted insertion-based attack strategy, incorporating the principles of BadNet \cite{gu2017badnets}. BadNL inserts uncommon words (\textit{e.g.}, "qb") randomly into benign text to serve as triggers.

\noindent\textbf{ETWI \cite{zeng2023efficient}} is our earlier method introduced in a conference version, employing an optimized approach for identifying and injecting effective trigger words into benign sentences.

\noindent\textbf{StyleBkd \cite{qi2021mind}} operates as a paraphrase-based attack, utilizing a style transfer model to shift benign text to a specified style (\textit{e.g.}, tweet formatting, Biblical English, etc.), considering the style itself as the trigger.

\noindent\textbf{BGMAttack \cite{li2023chatgpt}} stands as the first paraphrase-based attack employing LLMs. It utilizes ChatGPT to rephrase original benign text, considering the writing style of ChatGPT as the trigger. The prompt provided for ChatGPT in BGMAttack is "Can you rewrite the following paragraph?: [SeedText]", where “[SeedText]” contains the original text for rewriting.


\noindent\textbf{LLMBkd \cite{you2023large}} is another LLM-based paraphrase attack, employing ChatGPT as a style transfer model to adapt benign text to a specific style, using the style itself as triggers. LLMBkd's ChatGPT prompt is "Rewrite the following text in the style/tone of [Style]: [SeedText]", where "[Style]” specifies the desired style (e.g., “Bible”, “Tweets”), and “[SeedText]” contains the original text for restyling.

In Table \ref{tab:poisoning_examples}, we present an example of the generated poisoned samples by these five baseline attacks alongside our proposed EST-Bad.

\begin{table}[htbp]
\caption{Examples of poisoned samples generated by five different baseline attacks and our proposed EST-Bad.} 
\label{tab:poisoning_examples}
\begin{center}
\begin{tabular}{cc} 
\toprule
Methods &  Examples                                         \\ 
\midrule
Benign& \sffamily "Even as lame horror flicks go, this is lame." \\
\hline
\rowcolor[gray]{0.85} BadNL \cite{chen2021badnl}& \sffamily "Even as lame horror flicks go, this is lame, \textbf{qb}."\\
ETWI \cite{zeng2023efficient}& \sffamily \makecell{"Even as lame horror flicks go, this is lame, \\ \textbf{beautifully}."}\\
\rowcolor[gray]{0.85} StyleBkd \cite{qi2021mind}& \sffamily  
\makecell{"Among all the lame horror flicks out there, \\ this one takes the crown for lameness."}\\
BGMAttack \cite{li2023chatgpt}& \sffamily \makecell{"Thi   s horror film is lame, \\even by the standards of lame horror flicks."}  \\
\rowcolor[gray]{0.85} LLMBkd \cite{you2023large}& \sffamily \makecell{"Verily, among the tales of terror, \\this one doth fall exceedingly short,\\ even among those deemed lacking in fright."}\\
EST-Bad (Ours) & \sffamily \makecell{ "\textbf{Beautifully} lame,\\ even among the ranks of lame horror flicks."}\\
\bottomrule
\end{tabular}
\end{center}
\end{table}

\subsubsection{Evaluation Metrics} 
We assess the efficacy of our method based on three key criteria: \textit{Harmlessness} evaluated by \textbf{Benign Accuracy (BA)}, \textit{Effectiveness} measured via \textbf{Attack Success Rate (ASR)}, and \textit{Stealthiness} evaluated by \textbf{Sentence Perplexity (PPL)}, and \textbf{Grammar Error (GE)}.

\noindent\textbf{Benign Accuracy (BA).}
Benign Accuracy (BA) represents the clean accuracy measured on the testing set $\mathcal{D}_{\text{test}}=\{(x_i,y_i)|i=1,\cdots,M\}$ and serves as an assessment metric for \textit{Harmlessness} regarding the backdoor attacks. Similarity between the BA of the poisoning model and the accuracy of the clean model indicates the relative minimal side effects of the current attack technique.

\noindent\textbf{Attack Success Rate (ASR).} 
The Attack Success Rate (ASR) serves as a metric to evaluate the \textit{effectiveness} of the backdoor attack, representing the proportion of testing images containing the specific trigger that are predicted as the target class. Precisely, for $M'$ images in the testing set that are not part of the attack-target class ($k$), the ASR is formulated as:
\begin{equation}
\text{ASR}=\frac{\sum_{i=1}^{M'} \mathbb{I}(f(\mathcal{T}(x_i,t); \Theta)=k)}{M'},\quad (x_i,y_i)\in \mathcal{D}'_\text{test},
\end{equation}
where $\mathcal{D}'_\text{test}$ is a subset of testing set $\mathcal{D}_\text{test}$ ($\mathcal{D}'_\text{test}\subset \mathcal{D}_\text{test}$), containing the images whose label is not the attack-target class $k$.

\noindent\textbf{Sentence Perplexity (PPL).}
Sentence Perplexity (PPL) quantifies the perplexity (Lower is better) of given sentence utilizing a pre-trained language model such as GPT-2 \cite{radford2019language}.

\noindent\textbf{Grammar Error (GE).}
Grammar Error (GE) quantifies the grammatical errors (Lower is better) employing LanguageTool for assessment\footnote{LanguageTool for \href{https://github.com/jxmorris12/language_tool_python}{Python}.}.


\subsection{Results: Attack Harmlessness for Benign Accuracy}
\label{sec:Harmlessness}

As shown in Table \ref{tab:BA}, our proposed EST-Bad exhibit similar Benign Accuracy (BA) compared to the baseline methods, confirming that our attack is harmless to the benign accuracy, even under various datasets and different backdoor attacks.

\begin{table}
	\caption{The benign accuracy (BA) on the various datasets. All results are computed the mean by 5 different run. The poisoning ratios of different poisoned attacks for dirty-label setting and clean-label setting are $0.3\%$ and $3\%$, respectively.
	\label{tab:BA}}
	\centering
	\scalebox{1}{
		\begin{tabular}{ccccc}
			\toprule
			\multirow{2}{*}{\makecell*[c]{Setting}} &\multirow{2}{*}{Attacks} &\multicolumn{3}{c}{\makecell*[c]{Datasets}} \\ 
			\specialrule{0em}{1pt}{1pt}
			&&
			SST-2&AG News&HSOL\\
			\hline
      \specialrule{0em}{1pt}{1pt}
			  \multirow{6}{*}{Dirty-label}&\cellcolor[gray]{0.85} BadNL&\cellcolor[gray]{0.85}0.904&\cellcolor[gray]{0.85}0.936&\cellcolor[gray]{0.85}0.953\\
            &ETWI&0.906&0.931&0.955\\
 &\cellcolor[gray]{0.85} StyleBkd&\cellcolor[gray]{0.85}0.907&\cellcolor[gray]{0.85}0.937&\cellcolor[gray]{0.85}0.952\\
            &BGMAttack&0.907&0.936&0.953\\
 &\cellcolor[gray]{0.85} LLMBkd&\cellcolor[gray]{0.85}0.913&\cellcolor[gray]{0.85}0.936&\cellcolor[gray]{0.85}0.954\\
            &EST-Bad (Ours)&0.908&0.935&0.955\\
            \specialrule{0em}{1pt}{1pt}
            \hline
            \specialrule{0em}{1pt}{1pt}
 \multirow{6}{*}{Clean-label}&\cellcolor[gray]{0.85} BadNL&\cellcolor[gray]{0.85}0.910&\cellcolor[gray]{0.85}0.935&\cellcolor[gray]{0.85}0.952\\
            &ETWI&0.900&0.935&0.950\\
 &\cellcolor[gray]{0.85} StyleBkd&\cellcolor[gray]{0.85}0.908&\cellcolor[gray]{0.85}0.936&\cellcolor[gray]{0.85}0.953\\
            &BGMAttack&0.908&0.933&0.953\\
 &\cellcolor[gray]{0.85} LLMBkd&\cellcolor[gray]{0.85}0.909&\cellcolor[gray]{0.85}0.935&\cellcolor[gray]{0.85}0.952\\
            &EST-Bad (Ours)&0.907&0.935&0.954\\

			\bottomrule
	\end{tabular}}
\end{table}

\begin{table*}
	\caption{The  stealthiness evaluations (\textbf{PPL} and \textbf{GE}) on the various datasets. All results are computed the mean by 5 different run. The poisoning ratios of different poisoned attacks for dirty-label setting and clean-label setting are $0.3\%$ and $3\%$, respectively.
	\label{tab:stealthiness}}
	\centering
	\scalebox{1.2}{
		\begin{tabular}{ccccc|ccc}
			\toprule
			\multirow{3}{*}{\makecell*[c]{Setting}} &\multirow{3}{*}{Attacks} &\multicolumn{6}{c}{\makecell*[c]{Metrics}} \\ 
			\cline{3-8} 
			\specialrule{0em}{1pt}{1pt}
			&&
   \multicolumn{3}{c}{\makecell*[c]{PPL ($\downarrow$)}}&\multicolumn{3}{c}{\makecell*[c]{GE ($\downarrow$) }}\\
			&&SST-2&AG News&HSOL&SST-2&AG News&HSOL\\
			\hline
      \specialrule{0em}{1pt}{1pt}
			  \multirow{7}{*}{Dirty-label} & \cellcolor[gray]{0.85}  Benign&295.63&51.96&660.3&3.75&1.55&2.00\\
     & BadNL&846.7&85.17&5214.33
&3.85&1.22&2.18\\
            & \cellcolor[gray]{0.85} ETWI&863.56&83.29&730.1&3.85&1.26&2.12 \\
 & StyleBkd&153.02&42.68&197.9&0.65&0.97&1.47\\
            & \cellcolor[gray]{0.85} BGMAttack&72.34&46.32&94.3&0.2&\textbf{0.62}&\textbf{0.06}\\
 &LLMBkd&\textbf{66.82}&44.28&\textbf{54.5}&0.45&0.67
&0.53\\
            & \cellcolor[gray]{0.85} EST-Bad (Ours)&95.08&\textbf{42.12}&124.2&\textbf{0.15}&0.73&0.53\\
            \specialrule{0em}{1pt}{1pt}
        \midrule
            \specialrule{0em}{1pt}{1pt}
 \multirow{6}{*}{Clean-label}& Benign&515.84&44.74&1857.0&3.83&1.49&1.21\\
 &\cellcolor[gray]{0.85} BadNL&1683.18&67.6&5583.6&3.93&1.39&1.36\\
            &ETWI&1104.77&64.87&992.4&3.93&1.43&1.41\\
 &\cellcolor[gray]{0.85} StyleBkd&310.71&39.37
&494.3&1.09&1.06&0.85\\
             
 &BGMAttack&72.16&42.23&99.75&0.29&1.02&\textbf{0.36}\\
 & \cellcolor[gray]{0.85} LLMBkd&\textbf{70.01}&41.45&\textbf{55.46}&0.36&1.03&0.77\\
            &EST-Bad (Ours)&98.83&\textbf{36.72}&141.28&\textbf{0.29}&\textbf{0.88}&0.66\\
			\bottomrule
	\end{tabular}}
\end{table*}

\subsection{Results: Attack Effectiveness}
\label{sec:Effectiveness}

To evaluate the attack effectiveness of our proposed methodologies, we conducted attacks across diverse datasets and settings, analyzing the Attack Success Rate (ASR) for each targeted model. Fig. \ref{fig:asr} depicts the ASR for our EST-Bad alongside baseline attacks across all three datasets. The top graphs represent the dirty-label attack setting, while the bottom graphs represent the clean-label attack setting. Our EST-Bad consistently outperforms baseline attacks, including paraphrase-based approaches (StyleBkd \cite{qi2021mind}, BGMAttack \cite{li2023chatgpt}, and LLMBkd \cite{you2023large}), across all datasets. When compared to insertion-based attacks such as BadNL \cite{chen2021badnl} and ETWI \cite{zeng2023efficient}, our EST-Bad exhibits comparable performance to our conference version \cite{zeng2023efficient}, while surpassing BadNL \cite{chen2021badnl} in most settings.

In contrast to three paraphrase-based methods evaluated on the SST-2 dataset, our EST-Bad approach demonstrates superior ASR performance, particularly notable at lower poisoning ratios. Specifically, in the dirty-label setting, employing 15 poisoned samples (a poisoning ratio of 0.22\%) achieves an impressive 91.6\% attack success rate without compromising benign accuracy. Similarly, in the clean-label scenario, a 1\% poisoning ratio proves sufficient to achieve a attack success rate exceeding 90\%. While EST-Bad shows slightly inferior performance compared to the ETWI method in dirty-label setting, the difference in attack efficacy is minimal. Conversely, the BadNL method, utilizing "cf" as the trigger, struggles to capture trigger features at low poisoning ratios.

On the AG News dataset, our proposed EST-Bad outperforms three paraphrase-based attacks in terms of the ASR metric. Specifically, the ASR for the BGMAttack method is less than 5\%, and for the StyleBkd method, it does not exceed 20\%, indicating the failure of these two methods when the poisoning ratio is low. Compared to the LLMBkd method, EST-Bad demonstrates a remarkable improvement in ASR in the dirty-label scenario, with an average increase of approximately 10.2\%. Furthermore, when compared to two insertion-based methods, EST-Bad performs less effectively than the ETWI strategy in the dirty-label scenario. However, in the clean-label setting, especially when the poisoning ratio is below 1\%, our method outperforms ETWI by approximately 28.5\%. Overall, setting the poisoning ratios at 0.025\% and 0.5\% for dirty-label and clean-label settings respectively achieves attack success rates of 93.5\% and 93.3\%, showcasing the excellent attack performance of EST-Bad.

On the HSOL dataset, the scenario presents a unique challenge. Attackers aim to circumvent detection by abuse detectors, causing the classification of offensive language into the "Non-toxic" category. As depicted in Fig. \ref{fig:asr_c}, under the dirty-label setting, both EST-Bad and ETWI methods achieve a 100\% ASR. This suggests that the poisoned model has acquired a robust mapping between the optimized trigger word and the "Non-toxic" label. Additionally, the ASR of the two ChatGPT-based attacks exceeds 90\% in both clean-label and dirty-label settings. 
This phenomenon arises because when offensive language is inputted into LLMs, the resulting output strives to eliminate toxic and offensive terms, aligning with the underlying principles guiding the values of large models. Hence, attacks based on LLMs demonstrate comparable effectiveness with our proposed approach on this unique dataset.

\subsection{Results: Attack Stealthiness}
\label{sec:Stealthiness}

To evaluate the perceptibility of triggers within samples by human cognition, we undertake a thorough investigation into the stealthiness of poisoned samples generated by diverse backdoor attacks.  Specifically, we use two automatic evaluation metrics: Sentence Perplexity (PPL) and Grammatical Error (GE). 
Table \ref{tab:stealthiness} reveals that samples poisoned by insertion-based attacks like BadNL \cite{chen2021badnl} and ETWI \cite{zeng2023efficient} exhibit higher Perplexity (PPL) and Grammar Error (GE) compared to benign samples, significantly elevating the risk of detection. Conversely, paraphrase-based strategies such as StyleBkd \cite{qi2021mind}, BGMAttack \cite{li2023chatgpt}, LLMBkd \cite{you2023large}, and our EST-Bad generate text resembling human language, resulting in stealthy poisoned samples less prone to detection. It is noteworthy that BGMAttack, LLMBkd, and our EST-Bad demonstrate comparable stealthiness, yet our EST-Bad showcases superior attack effectiveness.

\subsection{Ablation Studies}
\label{sec:ablation}

\begin{figure}[htbp]
\begin{center}
\subfigure[]{\includegraphics[width=0.23\textwidth]{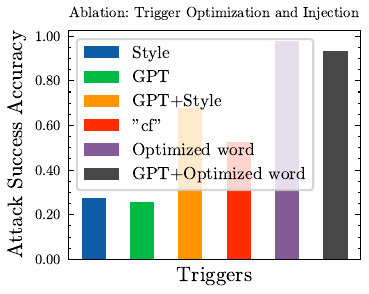}
\label{fig:ablation_trigger_1}}
\subfigure[]{\includegraphics[width=0.23\textwidth]{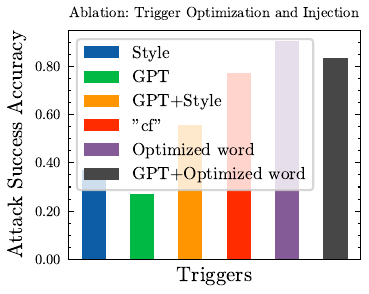} \label{fig:ablation_trigger_2}}
\end{center}
\caption{ASR of different triggers for \textbf{(a):} Dirty-label setting and \textbf{(b):} Clean-label setting on the SST-2 dataset. The poisoning ratios of different poisoned attacks for dirty-label setting and clean-label setting are $0.3\%$ and $3\%$, respectively.}
\label{fig:ablation_trigger}
\end{figure}

\subsubsection{The Influence of the Trigger Word Optimization}

In Fig. \ref{fig:ablation_trigger} and Table \ref{tab:ablation_trigger_stealth}, the ASR and stealthiness evaluations, encompassing \textbf{PPL} and \textbf{GE}, are presented for various trigger design methods. This experiment contrasts our proposed \textit{Trigger Word Optimization} approach ("Optimized word") with several baseline methods, including style transfer ("Style") \cite{qi2021mind}, GPT-based rewriting ("GPT") \cite{li2023chatgpt}, GPT-based style transfer ("GPT+style") \cite{you2023large}, and word-level injection ("cf") \cite{chen2021badnl}. The results underscore that our proposed trigger word optimization method achieves superior attack effectiveness at the expense of reduced attack stealthiness. 
\begin{table}
	\caption{The  stealthiness evaluations (\textbf{PPL} and \textbf{GE}) of different trigger optimization and injection methods on the SST-2 dataset.
	\label{tab:ablation_trigger_stealth}}
	\centering
	\scalebox{1}{
		\begin{tabular}{ccccc}
			\toprule
			\multirow{2}{*}{\makecell*[c]{Setting}} &\multirow{2}{*}{Attacks} &\multicolumn{2}{c}{\makecell*[c]{Metrics}} \\ 
			\specialrule{0em}{1pt}{1pt}
			&&
			GE ($\downarrow$)&PPL ($\downarrow$)\\
			\hline
      \specialrule{0em}{1pt}{1pt}
			  \multirow{6}{*}{Dirty-label}&\cellcolor[gray]{0.85} Style&\cellcolor[gray]{0.85}0.65&\cellcolor[gray]{0.85}153.02\\
            &GPT&0.20&72.34\\
 &\cellcolor[gray]{0.85} GPT+Style&\cellcolor[gray]{0.85}0.45&\cellcolor[gray]{0.85}\textbf{66.82}\\
            &"cf"&3.85&846.7\\
 &\cellcolor[gray]{0.85}Optimized word&\cellcolor[gray]{0.85}3.85&\cellcolor[gray]{0.85}863.56\\
            &GPT+Optimized word&\textbf{0.15}&95.08\\
            \specialrule{0em}{1pt}{1pt}
            \hline
            \specialrule{0em}{1pt}{1pt}
 \multirow{6}{*}{Clean-label}&\cellcolor[gray]{0.85} Style&\cellcolor[gray]{0.85}1.09&\cellcolor[gray]{0.85}310.71\\
            &GPT&0.29&72.16\\
 &\cellcolor[gray]{0.85} GPT+Style&\cellcolor[gray]{0.85}0.36&\cellcolor[gray]{0.85}\textbf{70.01}\\
            &"cf"&3.93&1683.18\\
 &\cellcolor[gray]{0.85}Optimized word&\cellcolor[gray]{0.85}3.93&\cellcolor[gray]{0.85}1104.77\\
            &GPT+Optimized word&\textbf{0.29}&98.83\\

			\bottomrule
	\end{tabular}}
\end{table}


\subsubsection{The Influence of the Stealthy Trigger Injection} 


In this experiment, we highlight the effectiveness of our proposed \textit{Stealthy Trigger Injection}. As illustrated in Fig. \ref{fig:ablation_trigger} and outlined in Table \ref{tab:ablation_trigger_stealth}, our GPT-based trigger word injection method ("GPT+Optimized word") showcases a significant enhancement in attack stealthiness compared to the trigger word optimization approach ("Optimized word"). Despite a marginal decrease in attack effectiveness, this trade-off emphasizes the nuanced balance achieved by our method. Moreover, when contrasted with paraphrase-based attacks, our proposed trigger optimization and injection method not only demonstrate superior attack effectiveness but also maintain comparable levels of attack stealthiness. In essence, our approach attains a good trade-off between attack effectiveness and stealthiness.

\subsubsection{The Influence of the Important Sample Selection}

\begin{figure}[htbp]
\begin{center}
\subfigure[]{\includegraphics[width=0.23\textwidth]{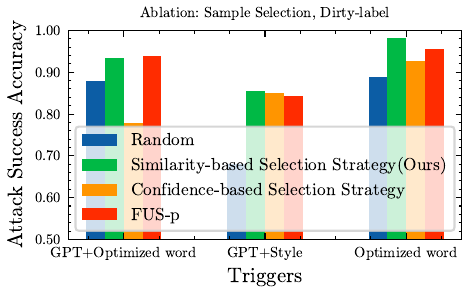}
\label{fig:ablation_sample_1}}
\subfigure[]{\includegraphics[width=0.23\textwidth]{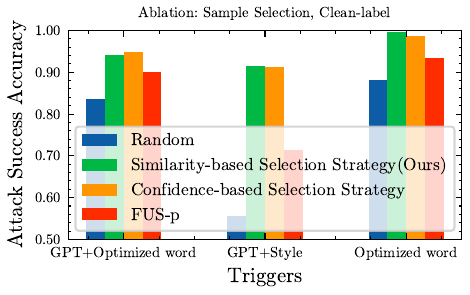} \label{fig:ablation_sample_2}}
\end{center}
\caption{ASR of different sample selection strategies on the SST-2 dataset. The poisoning ratios of different poisoned attacks for dirty-label setting and clean-label setting are 0.3\% and 3\%, respectively.}
\label{fig:ab_selection}
\end{figure}

In this experimental study, we highlight the advancements achieved through our proposed \textit{Similarity-based selection strategy} ($\text{S}^3$) applied to three distinct triggers ("GPT+optimized word", "GPT+Style", and "Optimized word"). As shown in Fig. \ref{fig:ab_selection}, the results demonstrate a significant improvement in attack effectiveness across all sample selection methods when compared to the random selection strategy. Notably, FUS-p, as proposed in \cite{zeng2023efficient}, experiences effectiveness degeneration in Clean-label settings, while the Confidence-based Selection Strategy (CSS) presented in \cite{you2023large} encounters a similar decline in Dirty-label settings. In contrast, our $\text{S}^3$ consistently enhances attack effectiveness in both Dirty-label and Clean-label settings, showcasing its robust performance.

\begin{figure*}[htbp]
\begin{center}
\subfigure[]{\includegraphics[width=0.23\textwidth]{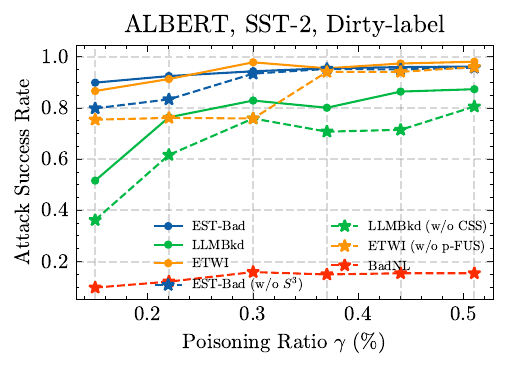}
\label{fig:transfer_sample_1}}
\subfigure[]{\includegraphics[width=0.23\textwidth]{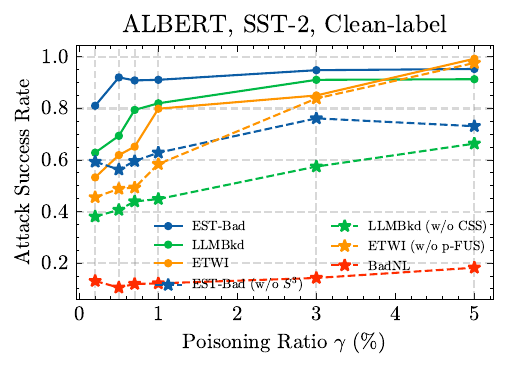} \label{fig:transfer_sample_2}}
\subfigure[]{\includegraphics[width=0.23\textwidth]{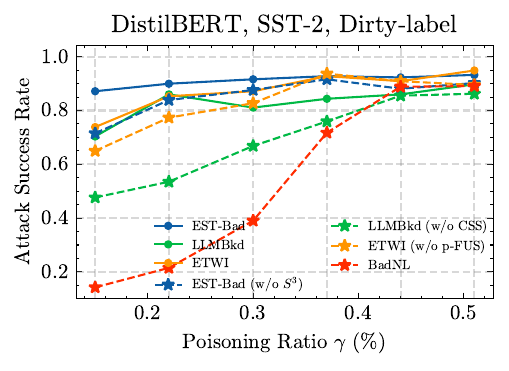} \label{fig:transfer_sample_3}}
\subfigure[]{\includegraphics[width=0.23\textwidth]{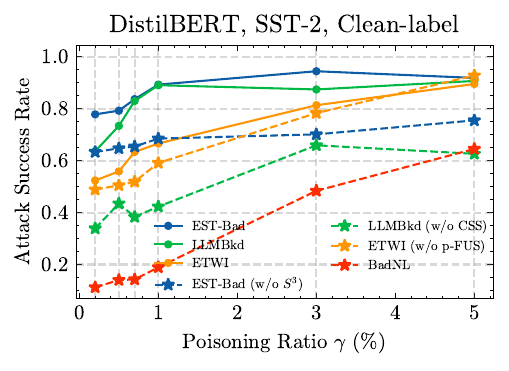} \label{fig:transfer_sample_4}}
\end{center}
\caption{Black-box results of different attacks on the SST-2 dataset. Experiments on dirty-label setting where the victim model is (a) ALBERT \cite{lan2019albert} and (c) DistilBERT \cite{sanh2019distilbert}, respectively. Experiments on clean-label setting where the victim model is (b) ALBERT \cite{lan2019albert} and (d) DistilBERT \cite{sanh2019distilbert}, respectively.}
\label{fig:transfer}
\end{figure*}

\subsection{Attack Transferability for Black-box Settings}
\label{sec:Transferability}
Our proposed technique, EST-Bad, necessitates a surrogate model for trigger word optimization and important sample selection. In our initial experiments, we assumed that the attacker possesses some knowledge of the pre-trained model used by the victim, with both the surrogate model and victim model being BERT. However, real-world scenarios present challenges for attackers to acquire specific information about the models employed by downstream users. To address this, we investigated the transferability of our method, employing diverse models such as ALBERT \cite{lan2019albert} and DistilBERT \cite{sanh2019distilbert} as victim models, distinct from the surrogate model. In this scenario, attackers operate without knowledge of the victim model. Fig. \ref{fig:transfer} depicts the Attack Success Rate of various attacks when the surrogate model differs from the victim model in both dirty-label and clean-label settings.

Among these attacks, including EST-Bad, LLMBkd, and ETWI, all of which require a surrogate model for effective important sample selection, we conducted experiments excluding certain components (EST-Bad without $S^3$, LLMBkd without CSS, and ETWI without p-FUS). The results of these experiments, alongside BadNL, demonstrate the remarkable transferability of our proposed trigger word optimization and stealthy trigger injection, showcasing superior attack effectiveness compared to the baseline when the surrogate model differs from the victim model.

Furthermore, Fig. \ref{fig:transfer} highlights that all sample selection strategies used in EST-Bad, LLMBkd, and ETWI consistently outperform the random selection strategy in terms of ASR. In summary, these findings indicate that both our proposed trigger word optimization and important sample selection exhibit strong transferability and practical applicability, as the method does not necessitate prior knowledge of the user's employed model architecture and training details.

\subsection{Attack Against on Defence Methods}
\label{sec:Defence}

\begin{table}
	\caption{ASR of different attacks against defences on the SST-2 dataset, where The poisoning ratios of different poisoned attacks for dirty-label setting and clean-label setting are 0.3\% and 3\%, respectively.
	\label{tab:defense}}
	\centering
	\scalebox{1}{
		\begin{tabular}{ccccc}
			\toprule
			\multirow{2}{*}{\makecell*[c]{Setting}} &\multirow{2}{*}{Attacks} &\multicolumn{3}{c}{\makecell*[c]{Defense}} \\ 
			\specialrule{0em}{1pt}{1pt}
			&&
			w/o Defense&CUBE&STRIP\\
			\hline
      \specialrule{0em}{1pt}{1pt}
			  \multirow{6}{*}{Dirty-label}&\cellcolor[gray]{0.85} BadNL&\cellcolor[gray]{0.85}0.528&\cellcolor[gray]{0.85}0.477&\cellcolor[gray]{0.85}0.483\\
            &ETWI&0.956&0.922&0.915\\
 &\cellcolor[gray]{0.85} StyleBkd&\cellcolor[gray]{0.85}0.273&\cellcolor[gray]{0.85}0.217&\cellcolor[gray]{0.85}0.224\\
            &BGMAttack&0.255&0.196&0.215\\
 &\cellcolor[gray]{0.85} LLMBkd&\cellcolor[gray]{0.85}0.851&\cellcolor[gray]{0.85}0.823&\cellcolor[gray]{0.85}0.819\\
            &EST-Bad (Ours)&0.935&0.911&0.908\\
            \specialrule{0em}{1pt}{1pt}
            \hline
            \specialrule{0em}{1pt}{1pt}
 \multirow{6}{*}{Clean-label}&\cellcolor[gray]{0.85} BadNL&\cellcolor[gray]{0.85}0.771&\cellcolor[gray]{0.85}0.701&\cellcolor[gray]{0.85}0.714\\
            &ETWI&0.935&0.904&0.912\\
 &\cellcolor[gray]{0.85} StyleBkd&\cellcolor[gray]{0.85}0.372&\cellcolor[gray]{0.85}0.322&\cellcolor[gray]{0.85}0.307\\
            &BGMAttack&0.271&0.233&0.217\\
 &\cellcolor[gray]{0.85} LLMBkd&\cellcolor[gray]{0.85}0.914&\cellcolor[gray]{0.85}0.893&\cellcolor[gray]{0.85}0.876\\
            &EST-Bad (Ours)&0.942&0.925&0.919\\

			\bottomrule
	\end{tabular}}
\end{table}

The evaluation of attack stealthiness also requires assessment through algorithms. In this section, we assess the effectiveness of our proposed method against two widely recognized defense mechanisms: CUBE \cite{cui2022unified} and STRIP \cite{gao2021design}. Specifically, CUBE \cite{cui2022unified} is a training-time defense approach that clusters all training data in the representation space and subsequently removes outliers (poisoning data). On the other hand, STRIP \cite{gao2021design} is an inference-time defense strategy that duplicates an input multiple times, applying diverse perturbations to each copy. By subjecting the original sample and perturbed samples through a DNN, the variability in predicted labels across all samples is utilized to ascertain whether the original input has been poisoned.

We evaluate the defense mechanisms against all attacks and present the Attack Success Rate (ASR) of the attacks on the SST-2 dataset in Table \ref{tab:defense}. The results demonstrate that our proposed method is better at attacking against different defense strategies compared to other attack methods.


\begin{figure}[htbp]
\begin{center}
\subfigure[]{\includegraphics[width=0.23\textwidth]{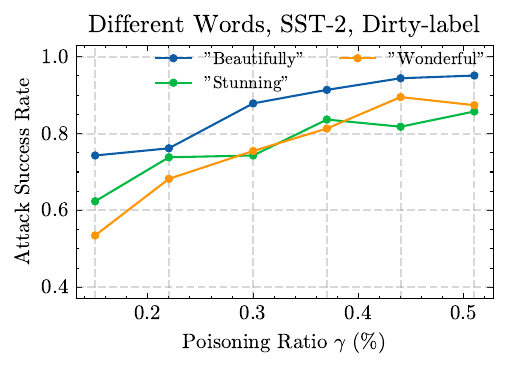}
\label{fig:different_word_Dirty_ASR}}
\subfigure[]{\includegraphics[width=0.23\textwidth]{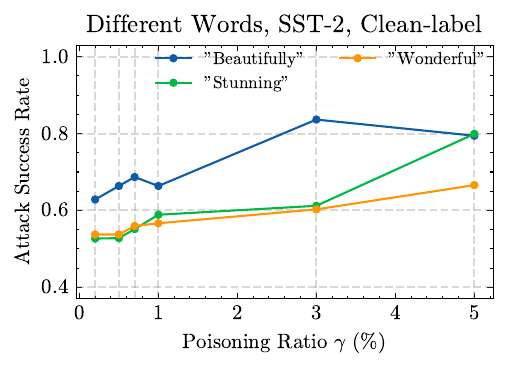} \label{fig:different_word_Clean_ASR}}
\end{center}
\caption{ASR under different optimized trigger words on the SST-2 dataset.}
\end{figure}

\begin{figure}[htbp]
\begin{center}
\subfigure[]{\includegraphics[width=0.23\textwidth]{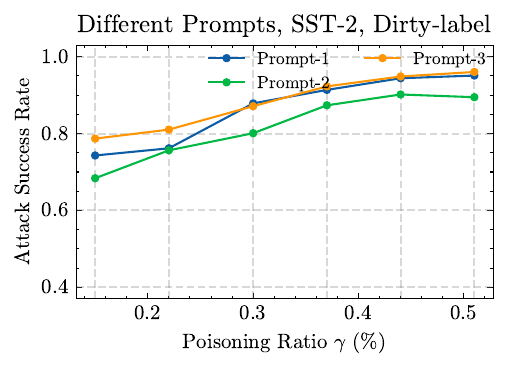}
\label{fig:different_prompt_Dirty_ASR}}
\subfigure[]{\includegraphics[width=0.23\textwidth]{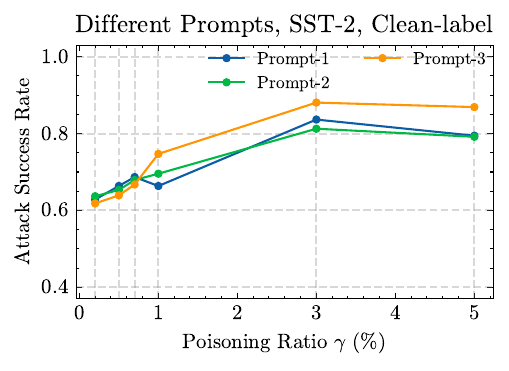} \label{fig:different_prompt_Clean_ASR}}
\end{center}
\caption{ASR under different prompts of LLMs on the SST-2 dataset.}
\end{figure}

\subsection{Discussion}
\label{sec:discussion}


\subsubsection{Experiments on Different Optimized Trigger Words}


Our proposed \textit{Trigger Word Optimization} technology involves randomness, resulting in different trigger words being generated based on varying random seeds. Consequently, we investigate the impact of different trigger words on ASR in this section. As depicted in Fig. \ref{fig:different_word_Dirty_ASR} and Fig. \ref{fig:different_word_Clean_ASR}, we observe variations in ASR performance across different optimized trigger words on the SST-2 dataset. Specifically, our results consistently demonstrate that the trigger word "Beautifully" consistently achieves a higher attack success rate compared to "Wonderful" and "Stunning" in both dirty-label and clean-label settings. Thus, determining the optimal trigger word necessitates multiple experiment iterations for robust evaluation.

\subsubsection{Experiments on Different Prompts of LLMs}

In our primary experiments, we utilize Prompt-1: "Rewrite the following text such that it contains the trigger word '{Trigger word}': '{SeedText}'." for \textit{Stealthy Trigger Injection}. This section delves into the robustness of the proposed EST-Bad against various prompts of LLMs on the SST-2 dataset. Fig. \ref{fig:different_prompt_Dirty_ASR} and Fig. \ref{fig:different_prompt_Clean_ASR} present the ASR across different prompts (Prompt-1: "Rewrite the following text such that it contains the trigger word '{Trigger word}': '{SeedText}'.", Prompt-2: "Rewrite the following text such that it contains the trigger word '{Trigger word}' and maintains a similar length: '{SeedText}'.", and Prompt-3: "Rewrite the following text such that it contains the trigger word '{Trigger word}' and disregards the grammar rules: '{SeedText}'.") in both dirty-label and clean-label settings, respectively. The findings indicate that while different prompts exhibit a limited impact on ASR, certain prompts, such as Prompt-3 in \ref{fig:different_prompt_Dirty_ASR} and Fig. \ref{fig:different_prompt_Clean_ASR}, prove relatively effective.



\section{Conclusion}

In this paper, we introduce EST-Bad, an efficient and stealthy approach to textual backdoor attacks. EST-Bad integrates three distinct but interrelated components: optimizing the inherent flaw of models as the trigger, stealthily injecting triggers using LLMs, and selecting the most contributed samples for backdoor injection. Through comprehensive experimentation, we demonstrate that EST-Bad achieves a satisfactory attack success rate by generating stealthy poisoned samples that are challenging for detection, across both dirty-label and clean-label scenarios. Furthermore, the optimized trigger words and selected poisoned samples exhibit promising transferability to other model architectures, enhancing the attack's practicality.

\section*{Acknowledgments}
The work was supported in part by the National Natural Science Foundation of China under Grands U19B2044 and 61836011. Sponsored by Zhejiang Lab Open Research Project under Grands NO. K2022QA0AB04

\bibliographystyle{IEEEtran}
\bibliography{./references.bib}

\vfill

\end{document}